%% file: main.tex
\documentclass[conference]{IEEEtran}
\IEEEoverridecommandlockouts

\usepackage{hyperref}
\usepackage{lipsum}  
\usepackage{cite}
\usepackage{amsmath,amssymb,amsfonts}
\usepackage{algorithmic}
\usepackage{graphicx}
\usepackage{textcomp}
\usepackage{booktabs}
\usepackage{multicol}
\usepackage{longtable}
\usepackage{array}
\usepackage{adjustbox}
\usepackage{caption}
\usepackage{tabularx, booktabs}
\usepackage[table, svgnames]{xcolor}

\usepackage{array}
\usepackage[multi-part-units=single]{siunitx}
\def\BibTeX{{\rm B\kern-.05em{\sc i\kern-.025em b}\kern-.08em
    T\kern-.1667em\lower.7ex\hbox{E}\kern-.125emX}}
\begin{document}

\title{A Passive Mechanical Add-on for Treadmill Exercise (P-MATE) in Stroke Rehabilitation}

\author{Irene L. Y. Beck\textsuperscript{1, 2,*}, Belle C. Hopmans\textsuperscript{1, 3}, Bram Haanen\textsuperscript{1}, Levi Kieft\textsuperscript{1},\\ Heike Vallery\textsuperscript{2, 3, 4}, Laura Marchal-Crespo\textsuperscript{1, 2}, and Katherine L. Poggensee\textsuperscript{1, 2, 3}
\thanks{This work was supported by the Health and Technology Convergence Alliance of TU Delft, Erasmus MC, and Erasmus University Rotterdam and the H2020-MSCA-RISE project CoSP- GA No 101106071 and Hocoma provided the Lokomat.}
\thanks{\textsuperscript{1}Dept. of Cognitive Robotics, TU Delft, Delft, Netherlands.}
\thanks{\textsuperscript{2}Dept. of Rehabilitation Medicine, Erasmus MC, Rotterdam, Netherlands.}
\thanks{\textsuperscript{3}Dept. of Biomechanical Engineering, TU Delft, Delft, Netherlands.}
\thanks{\textsuperscript{4}Institute of Automatic Control, RWTH Aachen University, Aachen, Germany.}
\thanks{Corresponding author: {\tt\small I.L.Y.Beck@tudelft.nl} }
}

\maketitle

\begin{abstract}
Robotic rehabilitation can deliver high-dose gait therapy and improve motor function after a stroke. However, for many devices, high costs and lengthy setup times limit clinical adoption. Thus, we designed, built, and evaluated the Passive Mechanical Add-on for Treadmill Exercise (P-MATE), a low-cost passive end-effector add-on for treadmills that couples the movement of the paretic and non-paretic legs via a reciprocating system of elastic cables and pulleys. Two human-device mechanical interfaces were designed to attach the elastic cables to the user. The P-MATE and two interface prototypes were tested with a physical therapist and eight unimpaired participants. Biomechanical data, including kinematics and interaction forces, were collected alongside standardized questionnaires to assess usability and user experience. Both interfaces were quick and easy to attach, though user experience differed, highlighting the need for personalization. We also identified areas for future improvement, including pretension adjustments, tendon derailing prevention, and understanding long-term impacts on user gait. Our preliminary findings underline the potential of the P-MATE to provide effective, accessible, and sustainable stroke gait rehabilitation.
\end{abstract}


\input{01_Introduction}
\input{02_Design}

\input{03_Experiment}
\input{04_Discussion}
\input{04b_Conclusion}
\bibliographystyle{IEEEtran}
\bibliography{05_Bibliography}

\end{document}

%% file: 01_Introduction.tex
\section{Introduction} 
Stroke is one of the leading causes of disability worldwide, affecting twelve million people annually~\cite{Feigin_2022}. Up to \SI{75}{\%} of people who have had a stroke are left with lower-limb somatosensory and motor impairments three months post-stroke~\cite{Dobkin_2005}. Individuals with more severe impairments experience a greater negative impact on their quality of life and participation in everyday activities~\cite{Dobkin_2005}.

High-dose physical therapy can enhance motor recovery following a stroke~\cite{Tollar_2021}. However, individuals often receive less therapy than recommended due to, among other factors, 
a shortage of trained personnel~\cite{Foley_N_2012}. Cnventional gait interventions---where therapists physically assist or resist an individual's limb movements---usually require more than one therapist per patient~\cite{Dobkin_2005}. These physical demands placed on therapists can lead to work-related injuries~\cite{Foley_S_2024}. These shortcomings in current lower-limb rehabilitation practices are expected to worsen with an aging population and global physician shortage~\cite{Haakenstad_2022}.
\begin{figure}[t!]
    \centering
    \def\svgwidth{1\linewidth}
    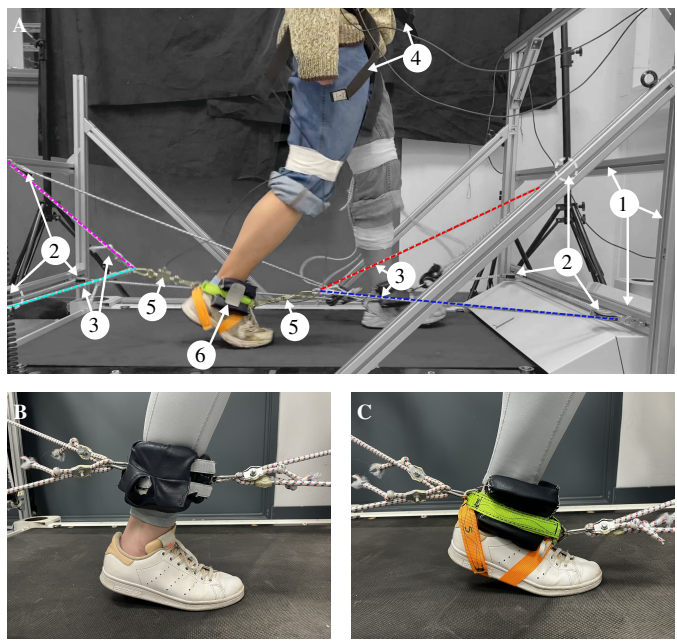
    \caption{The mechanical design of the P-MATE. \textbf{A}: An overview of all parts of the system with the 1) frame, 2) six pulleys, 3) two anterior and two posterior elastic tendons, 4) body weight support, 5) anterior and posterior load sensors, and 6) human-device interface. The two prototypes of the interface are highlighted in \textbf{B}: the cuff around the shank and \textbf{C}: the bandage around the foot.}
    \label{fig:working_principle}
\end{figure} 

Robotic rehabilitation provides a promising solution to these challenges by delivering high-intensity therapy while reducing physical demands on therapists. Several robotic devices have already been developed to support gait rehabilitation, offering controlled, high-intensity training as a way to lower therapists' workload while objectively tracking patient progress~\cite{Pennycott_2012, Langhorne_2011}. These devices can be categorized as end-effector robots and exoskeletons. End-effector robots interact with the body at a single point, typically the foot. In contrast, exoskeletons often assist at multiple joints and require attachment to several body parts, which increases the time needed to put them on.  Exoskeletons are often used in combination with a treadmill and (partial) body weight support (BWS) systems~\cite{Pennycott_2012}. 

However, factors such as set-up time and cost hinder the introduction and adoption of robotics into the clinic, calling for more affordable and simpler devices~\cite{Foley_S_2024}. Despite advances in the design of simpler lightweight exoskeletons~\cite{Awad_2017}, current robotic solutions can limit the active participation of patients, which is crucial to enhance neurorehabilitation~\cite{Miguel-Fernandez_2023, Pennycott_2012}.

Recent research showed the potential of tendon-based gait rehabilitation devices~\cite{Hidayah_2020, Foley_S_2024, Jin_2018}. Yet, how to route the cables to effect assistance in the impaired leg without impeding the unimpaired leg's movement and range of motion is still an open question.
One option is to attach a single tendon to the impaired leg, which tensions a spring in a controlled way during the stance phase of the impaired leg and subsequently releases the stored energy during swing~\cite{Tomc_2022}. However, the mechanism still requires an active brake, electronics, and control to compensate for the movement of the impaired leg during stance and to avoid tensioning of the spring during the entire stance phase.

We aimed to address the aforementioned challenges by designing and evaluating a purely mechanical solution that exploits the reciprocating movement of both legs to assist in the gait rehabilitation of the paretic leg. This is done by attaching both legs to elastic tendons, guided by a set of pulleys optimally located around a treadmill to leverage the movement of the non-paretic leg and the treadmill. This low-cost system could be implemented on existing treadmills, does not require any actuators or control, and allows for the individual's active participation. 

%% file: F1_ILYB_2.pdf_tex
\begingroup%
  \makeatletter%
  \providecommand\color[2][]{%
    \errmessage{(Inkscape) Color is used for the text in Inkscape, but the package 'color.sty' is not loaded}%
    \renewcommand\color[2][]{}%
  }%
  \providecommand\transparent[1]{%
    \errmessage{(Inkscape) Transparency is used (non-zero) for the text in Inkscape, but the package 'transparent.sty' is not loaded}%
    \renewcommand\transparent[1]{}%
  }%
  \providecommand\rotatebox[2]{#2}%
  \newcommand*\fsize{\dimexpr\f@size pt\relax}%
  \newcommand*\lineheight[1]{\fontsize{\fsize}{#1\fsize}\selectfont}%
  \ifx\svgwidth\undefined%
    \setlength{\unitlength}{1218.8976378bp}%
    \ifx\svgscale\undefined%
      \relax%
    \else%
      \setlength{\unitlength}{\unitlength * \real{\svgscale}}%
    \fi%
  \else%
    \setlength{\unitlength}{\svgwidth}%
  \fi%
  \global\let\svgwidth\undefined%
  \global\let\svgscale\undefined%
  \makeatother%
  \begin{picture}(1,0.94673001)%
    \lineheight{1}%
    \setlength\tabcolsep{0pt}%
    \put(0,0){\includegraphics[width=\unitlength,page=1]{F1_ILYB_2.pdf}}%
    \put(0.21534585,0.48735156){\color[rgb]{0,0,0}\makebox(0,0)[t]{\lineheight{1.10000002}\smash{\begin{tabular}[t]{c}\footnotesize{5}\end{tabular}}}}%
    \put(0,0){\includegraphics[width=\unitlength,page=2]{F1_ILYB_2.pdf}}%
    \put(0.28891669,0.41530779){\color[rgb]{0,0,0}\makebox(0,0)[t]{\lineheight{1.10000002}\smash{\begin{tabular}[t]{c}\footnotesize{6}\end{tabular}}}}%
    \put(0,0){\includegraphics[width=\unitlength,page=3]{F1_ILYB_2.pdf}}%
    \put(0.42926721,0.45735733){\color[rgb]{0,0,0}\makebox(0,0)[t]{\lineheight{1.10000002}\smash{\begin{tabular}[t]{c}\footnotesize{5}\end{tabular}}}}%
    \put(0,0){\includegraphics[width=\unitlength,page=4]{F1_ILYB_2.pdf}}%
    \put(0.58897393,0.53362352){\color[rgb]{0,0,0}\makebox(0,0)[t]{\lineheight{1.10000002}\smash{\begin{tabular}[t]{c}\footnotesize{3}\end{tabular}}}}%
    \put(0,0){\includegraphics[width=\unitlength,page=5]{F1_ILYB_2.pdf}}%
    \put(0.8394818,0.55443254){\color[rgb]{0,0,0}\makebox(0,0)[t]{\lineheight{1.10000002}\smash{\begin{tabular}[t]{c}\footnotesize{2}\end{tabular}}}}%
    \put(0,0){\includegraphics[width=\unitlength,page=6]{F1_ILYB_2.pdf}}%
    \put(0.92351812,0.64247043){\color[rgb]{0,0,0}\makebox(0,0)[t]{\lineheight{1.10000002}\smash{\begin{tabular}[t]{c}\footnotesize{1}\end{tabular}}}}%
    \put(0,0){\includegraphics[width=\unitlength,page=7]{F1_ILYB_2.pdf}}%
    \put(0.61298431,0.86096465){\color[rgb]{0,0,0}\makebox(0,0)[t]{\lineheight{1.10000002}\smash{\begin{tabular}[t]{c}\footnotesize{4}\end{tabular}}}}%
    \put(0,0){\includegraphics[width=\unitlength,page=8]{F1_ILYB_2.pdf}}%
    \put(0.06990206,0.57167504){\color[rgb]{0,0,0}\makebox(0,0)[t]{\lineheight{1.10000002}\smash{\begin{tabular}[t]{c}\footnotesize{2}\end{tabular}}}}%
    \put(0.02986235,0.91072078){\color[rgb]{1,1,1}\makebox(0,0)[rt]{\lineheight{1.10000002}\smash{\begin{tabular}[t]{r}\footnotesize{\textbf{A}}\end{tabular}}}}%
    \put(0.54693675,0.32966791){\color[rgb]{1,1,1}\makebox(0,0)[rt]{\lineheight{1.10000002}\smash{\begin{tabular}[t]{r}\footnotesize{\textbf{C}}\end{tabular}}}}%
    \put(0.02986236,0.32966791){\color[rgb]{1,1,1}\makebox(0,0)[rt]{\lineheight{1.10000002}\smash{\begin{tabular}[t]{r}\footnotesize{\textbf{B}}\end{tabular}}}}%
    \put(0,0){\includegraphics[width=\unitlength,page=9]{F1_ILYB_2.pdf}}%
    \put(0.1293246,0.45973784){\color[rgb]{0,0,0}\makebox(0,0)[t]{\lineheight{1.10000002}\smash{\begin{tabular}[t]{c}\footnotesize{3}\end{tabular}}}}%
  \end{picture}%
\endgroup%

%% file: 02_Design.tex
\section{Design} 

\subsection{System Overview}
We developed the Passive Mechanical Add-on for Treadmill Exercise (P-MATE) for use in stroke gait rehabilitation. The P-MATE was designed to provide support during treadmill walking, particularly for individuals with hemiparesis, a condition that impairs gait symmetry by weakening one side of the body. The add-on combines four elastic tendons attached to both legs and guided by pulleys. The pulleys are optimally located around a treadmill to assist the paretic leg as needed during the swing phase while the non-paretic leg is moved backward by the treadmill during stance, but without impeding physiological leg motion. 
A schematic overview of the P-MATE's working principle is presented in Fig.~\ref{fig:working_principle}A.

Two human-device interfaces were designed to attach the elastic tendons to different points on the user: one at the shank (\textit{cuff prototype}) and the other at the foot (\textit{bandage prototype}), see Figs.~\ref{fig:working_principle}B and~\ref{fig:working_principle}C, respectively. 
The entire system was designed and evaluated via two datasets, both of which were approved by the Human Research Ethics Committee of Delft University of Technology, and all participants provided written informed consent prior to the start of the study.

\subsection{Pulley Location Optimization}
\begin{figure}[t]
    \centering
    \def\svgwidth{1\linewidth}
    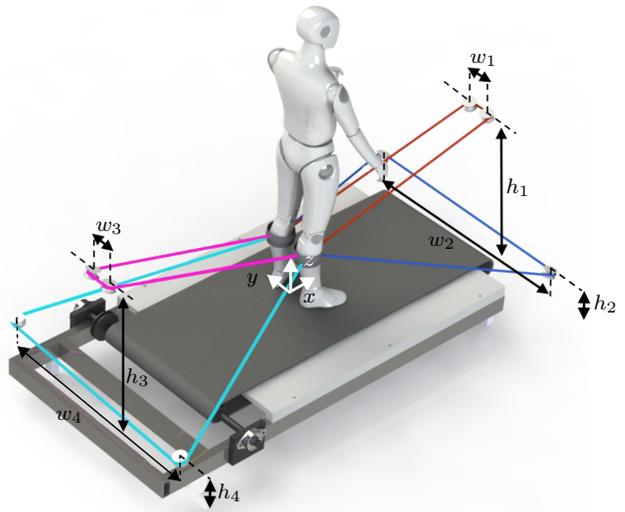
    \caption{The optimization setup with the eight parameters to determine optimal pulley location.}
    \label{fig:opt_parameters}
\end{figure}

The goal of the P-MATE was to assist the paretic foot’s movement in the three translational degrees of freedom (DoFs), with coordinate directions forward ($x$), lateral ($y$), and upward ($z$) (Fig.~\ref{fig:opt_parameters}). Therefore, a minimum of four elastic tendons is needed due to the underconstrained behavior of tendons (i.e., only pulling)~\cite{Fang_2005}. Furthermore, to promote gait symmetry, the pulley and tendon arrangement is symmetrical across the sagittal plane. 

The P-MATE was designed to avoid interfering with unimpaired gait. To achieve this, a simulation-based optimization was run to determine the optimal pulley locations that minimize the ``parasitic'' forces, i.e., any forces acting on the legs during ``normal gait'' for six participants. The optimization used a cost function that minimized the sum of the root mean square of each of the Cartesian forces (in $x$-, $y$- and $z$-directions) acting on the left and right legs over time. At each frame (collected at \SI{100}{Hz}), the Cartesian forces were derived from the ``parasitic'' tendon forces, which were computed using a Jacobian based on the location of the legs and the pulleys.  

These tendon ``parasitic'' forces originate from the elongation of the elastic tendons during unimpaired walking. Based on preliminary research of commercially-available elastic ropes, we chose to set all tendons to the same stiffness of \SI{300}{N/m}. The slack length, or the length at zero-force, of each tendon was chosen as the shortest tendon length observed per participant. 

The optimization parameters describe the locations of eight pulleys around the treadmill, two pulleys per each of the four tendons, while a simulated human was placed in the center of the treadmill. The pulley locations were assumed constant in the $x$-direction at \SI{0.75}{m} in front of and behind the individual, based on conventional treadmill length. Additionally, the location of each pair of pulleys was assumed to be the same height. These assumptions reduce the optimization parameters to only the distances between the pulleys per pair in $y$-direction ($w_{j}$) and their heights ($h_{j}$) (see Fig.~\ref{fig:opt_parameters}), resulting in four widths and four heights.
The search space of the parameters was constrained to $w_{1,3} \in [0, 0.1]$, $w_{2,4} \in [0.4, 0.8]$, $h_{1,3} \in [0.5, 1.0]$, and $h_{2,4} \in [0.1, 0.3]$~\SI{}{m}. Initial conditions were set to the average of each constraint bound. 
The optimization of the pulleys was executed for data collected from six unimpaired young adults walking on a treadmill at \SI{1.0}{m/s} (see Table~\ref{tab:opt_subjects}). We used the genetic algorithm (\verb|ga|) from the Optimization Toolbox in MATLAB (R2023b, Mathworks, USA). The optimization stopped after \SI{100}{} generations or when the average relative change in the cost function was smaller than \SI{1e-6}{N}.   

\begin{table}
\centering
\caption{Demographic information for pulley location optimization training dataset}
\begin{tabular}{lllll}
\toprule
\textbf{ID} &\textbf{Sex}  & \textbf{Age/\si{yr}}  & \textbf{Height/\si{m}} & \textbf{Mass/\si{kg}}   \\ \midrule
A      & F        & 30 & 1.55          & 51.0                      \\ 
B      & M          & 26 & 1.73          & 63.7                      \\ 
C      & M          & 25 & 1.76          & 63.6                      \\ 
D      & F        & 23 & 1.71          & 56.2                      \\
E      & M          & 30 & 1.87          & 72.5                      \\
F      & M          & 30 & 1.81          & 71.0                     \\
\bottomrule
\end{tabular}
\label{tab:opt_subjects}
\end{table}

The main contributors to the cost function were the forces in $x$-direction, with median anterior and posterior forces of \SI{28.4}{N} and \SI{34.0}{N}, respectively. The optimal pulley location parameters per participant are shown in Table~\ref{tab:opt_pulley_loc}. The optimal solution for both $w_1$ and $w_3$, i.e., the width between the two sets of upper pulleys, was \SI{0.00}{m} for all participants, reducing the overall system from eight to six pulleys.

\input{0T_Optimal_pulley_ILYB}
\subsection{Mechanical Design}
\subsubsection{Frame}
A frame was built to fit around a modified Lokomat (Hocoma AG, Switzerland) treadmill to realize P-MATE based on the optimization results. The frame consists of extruded aluminum Profile 8 40x40 profiles (item, Germany) of various lengths anchored to the treadmill frame. The total dimensions of the P-MATE are \SI{1.8}{m} long by \SI{1.0}{m} wide by \SI{1.2}{m} high. 

\subsubsection{Pulleys}
Based on the optimal solutions found in the simulation, six pulleys were used (diameter \SI{50}{mm}, groove \SI{6}{mm}) (105140, Waelbers, Netherlands). 
Three pulleys are attached to the front part of the frame, and three pulleys to the back, with one higher pulley in the middle with two pulleys at treadmill height on either side. 
The implemented pulley locations (Imp in Table~\ref{tab:opt_pulley_loc}) were kept as close to the optimal solution subject to the setup constraints. This includes a slight increase in the height of the lower lateral pulleys and lowering of the upper back pulley, while the $x$-position of all pulleys was increased to \SI{+-0.}{m}. The layout of the pulleys is symmetrical in the sagittal and coronal planes.
\subsubsection{Elastic tendons}
ons are commercially available with a diameter of \SI{8}{mm} (5411387036028, Ledent, Corbeo NV, Belgium), chosen in an effort to match the assumptions from the simulation. Four stainless steel carabiners (432712, Seilflechter, Germany) are connect the elastic tendons to the human-device interface. The slack length of the elastic tendons can be adjusted via a rope tightener for each elastic tendon (CAMJAM\texttrademark XT\texttrademark, Nite Ize, USA).

\subsubsection{Safety harness}
The harness of a BWS system is used to provide safety to the individual against falls. There was enough slack in the system to allow for free movement. However, any safety/BWS system could be used in combination with the P-MATE.

\subsubsection{Human-device interface}
The P-MATE requires a human-device interface to effectively transfer the elastic tendons' forces to the user while promoting comfort, ease of use, and adjustability. To identify user requirements for the interface, observations of training sessions and interviews with a neurorehabilitation physical therapist (\SI{27}{years} experience) were conducted at a stroke rehabilitation center (Rijndam Revalidatie, the Netherlands). 
These activities indicated a need for a quick and easy setup for rehabilitation devices to maximize the therapy time. This includes not only a maximum of five minutes for the donning and doffing of the system on the patient but also reduced complexity compared to existing robotic exoskeletons. Additionally, the system should accommodate patients typically seen in the clinic, ranging in demographics and severity of disease. We thereforecuffbandage.     
The cuff prototype encloses the shank via an aluminum C-shaped profile around the back of the shank and a rectangular aluminum plate at the front (Fig.~\ref{fig:working_principle}B). The aluminum is covered by a layer of cold foam padding with an exterior layer of synthetic leather to allow for a comfortable but close-fitting fit to the subject's leg. 
The bandage prototype (see Fig.~\ref{fig:working_principle}C) consists of two straps, one around the ankle and one in a figure-eight pattern wrapping around the sole of the foot. To prevent shearing between the strap and the ankle, cold foam padding covered in synthetic leather was added to the ankle strap. 

The amount of Velcro used to secure both prototypes to the participant was chosen to withstand the worst-case scenario, i.e., where the participant was at the limits of the treadmill and the pulling forces were the highest. To facilitate donning and doffing, stainless steel buckles at the front and back of the interfaces secure the elastic tendons using the carabiners.

%% file: F2_v2.pdf_tex
\begingroup%
  \makeatletter%
  \providecommand\color[2][]{%
    \errmessage{(Inkscape) Color is used for the text in Inkscape, but the package 'color.sty' is not loaded}%
    \renewcommand\color[2][]{}%
  }%
  \providecommand\transparent[1]{%
    \errmessage{(Inkscape) Transparency is used (non-zero) for the text in Inkscape, but the package 'transparent.sty' is not loaded}%
    \renewcommand\transparent[1]{}%
  }%
  \providecommand\rotatebox[2]{#2}%
  \newcommand*\fsize{\dimexpr\f@size pt\relax}%
  \newcommand*\lineheight[1]{\fontsize{\fsize}{#1\fsize}\selectfont}%
  \ifx\svgwidth\undefined%
    \setlength{\unitlength}{1359.54510498bp}%
    \ifx\svgscale\undefined%
      \relax%
    \else%
      \setlength{\unitlength}{\unitlength * \real{\svgscale}}%
    \fi%
  \else%
    \setlength{\unitlength}{\svgwidth}%
  \fi%
  \global\let\svgwidth\undefined%
  \global\let\svgscale\undefined%
  \makeatother%
  \begin{picture}(1,0.76569729)%
    \lineheight{1}%
    \setlength\tabcolsep{0pt}%
    \put(0,0){\includegraphics[width=\unitlength,page=1]{F2_v2.pdf}}%
    \put(0.75328121,0.67765639){\color[rgb]{0,0,0}\makebox(0,0)[t]{\lineheight{0.80000001}\smash{\begin{tabular}[t]{c}\footnotesize{$w_1$}\end{tabular}}}}%
    \put(0.19142814,0.42429348){\color[rgb]{0,0,0}\makebox(0,0)[t]{\lineheight{0.80000001}\smash{\begin{tabular}[t]{c}\footnotesize{$w_3$}\end{tabular}}}}%
    \put(0.13126852,0.14141799){\color[rgb]{0,0,0}\makebox(0,0)[t]{\lineheight{0.80000001}\smash{\begin{tabular}[t]{c}\footnotesize{$w_4$}\end{tabular}}}}%
    \put(0.6903996,0.40257668){\color[rgb]{0,0,0}\makebox(0,0)[t]{\lineheight{0.80000001}\smash{\begin{tabular}[t]{c}\footnotesize{$w_2$}\end{tabular}}}}%
    \put(0.37180419,0.023218){\color[rgb]{0,0,0}\makebox(0,0)[t]{\lineheight{0.80000001}\smash{\begin{tabular}[t]{c}\footnotesize{$h_4$}\end{tabular}}}}%
    \put(0.80492527,0.48201458){\color[rgb]{0,0,0}\makebox(0,0)[t]{\lineheight{0.80000001}\smash{\begin{tabular}[t]{c}\footnotesize{$h_1$}\end{tabular}}}}%
    \put(0.93627245,0.3063996){\color[rgb]{0,0,0}\makebox(0,0)[t]{\lineheight{0.80000001}\smash{\begin{tabular}[t]{c}\footnotesize{$h_2$}\end{tabular}}}}%
    \put(0.23367715,0.19615467){\color[rgb]{0,0,0}\makebox(0,0)[t]{\lineheight{0.80000001}\smash{\begin{tabular}[t]{c}\footnotesize{$h_3$}\end{tabular}}}}%
    \put(0.49113419,0.37314478){\color[rgb]{1,1,1}\makebox(0,0)[t]{\lineheight{0.80000001}\smash{\begin{tabular}[t]{c}\footnotesize{$z$}\end{tabular}}}}%
    \put(0.40619348,0.34694214){\color[rgb]{0,0,0}\makebox(0,0)[t]{\lineheight{0.80000001}\smash{\begin{tabular}[t]{c}\footnotesize{$y$}\end{tabular}}}}%
    \put(0.4917014,0.31800711){\color[rgb]{0,0,0}\makebox(0,0)[t]{\lineheight{0.80000001}\smash{\begin{tabular}[t]{c}\footnotesize{$x$}\end{tabular}}}}%
  \end{picture}%
\endgroup%

%% file: 0T_Optimal_pulley_ILYB.tex
\begin{table}[t]
\centering
\caption{The optimal pulley locations for each participant from the simulation-based optimization and the final implemented values (Imp) in the mechanical design. All measurements are in meters. Imp: Implemented.}
\begin{tabular}{lllllllll}
\toprule
\textbf{ID}& $w_{1}$    & $h_{1}$       & $w_{2}$ & $h_{2}$          & $w_{3}$   & $h_{3}$     & $w_{4}$     & $h_{4}$    \\ \midrule
A                   & 0.00       & 0.50          & 0.40        & 0.10          & 0.00       & 0.74        & 0.56        & 0.10       \\
B                   & 0.00       & 0.50          & 0.40        & 0.10          & 0.00       & 0.76        & 0.40        & 0.10       \\
C                   & 0.00       & 0.57          & 0.80        & 0.10          & 0.00       & 0.69        & 0.80        & 0.10       \\
D                   & 0.00       & 0.50          & 0.74        & 0.10          & 0.00       & 0.62        & 0.40        & 0.10       \\
E                   & 0.00       & 0.56          & 0.40        & 0.10          & 0.00       & 0.78        & 0.80        & 0.10       \\
F                   & 0.00       & 0.56          & 0.40        & 0.10          & 0.00       & 0.79        & 0.55        & 0.10       \\
Imp         & 0.00       & 0.50          & 0.80        & 0.13          & 0.00       & 0.50        & 0.80        & 0.13       \\
\bottomrule
\end{tabular}%
\label{tab:opt_pulley_loc}
\end{table}

%% file: 03_Experiment.tex
\section{Experiment} 
\subsection{Experimental Protocol}
We conducted a feasibility study with eight adults without physical impairments (see Table \ref{tab:exp_subjects}) to assess TE and the two interface prototypes.

Each participant received a demonstration from the experimenter of how to don and doff the two prototypes while sitting in a comfortable chair. They were then invited to walk on the treadmill and select their comfortable walking speed and pretension. They were then asked to walk at their selected speed for five minutes while wearing the safety harness but without any prototype (baseline). Following the five-minute baseline, participants took a minimum five-minute rest while sitting on a comfortable chair, during which the experimenter donned one of the prototypes.

Participants then walked for \SI{20}{minutes} at their chosen speed with each prototype, in a randomized order. After each session, participants completed three questionnaires to evaluate the prototype's design while resting for at least five minutes. During this break, the experimenter doffed the first prototype and donned the second, timing each transition.

After both walking sessions, participants practiced donning and doffing each prototype independently. The experimenter timed these actions, and participants completed the final section of the questionnaires.

The neurorehabilitation physical therapist, 
who contributed to identifying user requirements, also participated in a shortened experimental protocol. Baseline walking measurements were taken for one minute, followed by ten-minute walking sessions with each prototype. At the therapist's request, a second ten-minute session with the bandage prototype was conducted with lengthened anterior elastic tendons.

\subsection{Outcome metrics}
During the baseline and the two walking sessions, biomechanical data regarding the right leg were collected. In particular, forces at the front and back of the leg were recorded with two load sensors (KD40S 1000N, ME-Me{\ss}systeme, Germany). One load sensor was positioned between the two anterior tendons and the human-device interface on the front of the right leg, while a second one was placed between the two posterior elastic tendons and the same interface on the right leg.

To evaluate whether P-MATE changes normal walking, a depth camera (ZED mini Stereo Camera, StereoLabs, USA) with the onboard software detecting anatomical landmarks was used to capture the kinematics of the right leg in the sagittal plane. Step length and step height were extracted from this kinematic data. The force and kinematic data were synchronized offline. 

To evaluate setup time, the experimenter timed every donning and doffing action, either performed by the experimenter herself or the participant.  
Data on participants' user experience and acceptance of the prototypes, and system usability were collected using three questionnaires: the Van Der Laan Acceptance Questionnaire \cite{vanderLaan_1997}, the System Usability Scale (SUS) \cite{Brooke_1995}, and a modified version of the User Experience Questionnaire (UEQ) \cite{Laugwitz_2008}, respectively. The questions in the UEQ were modified to target the experience with the P-MATE. The Van Der Laan Acceptance Questionnaire and the UEQ were answered on a Likert scale from -2 to +2, with -2 indicating a negative response (e.g., ``useless'') and +2 a positive response (e.g., ``useful''). The SUS was answered on a Likert scale from 1 to 5: 1 indicating ``Strongly disagree'' and 5 ``Strongly agree,'' resulting in an overall score between 0 and 100. Each questionnaire was filled out for both interface prototypes. The experimenter also wrote down any comments made by the participants during the study.

\subsection{Experimental Results}
\input{0T_Questionnaire}
For both prototypes, the highest anterior tendon forces occurred at toe off (determined as the instance the foot was farthest back), while the highest posterior tendon forces were recorded at heel strike (conversely the instance the foot was farthest forward). In the cuff prototype, the highest average peak anterior and posterior forces across all participants were \SI{118}{N} and \SI{110}{N}, respectively. For the bandage prototype, these forces reached \SI{141}{N} and \SI{114}{N}, respectively.

All participants showed larger differences between the maximum and minimum forces with the bandage prototype compared to the cuff prototype. Over time, comparing minutes 2 and 19 (early and late in the session), the cuff prototype visually appeared to exert consistent forces at the front and back of the leg, whereas the forces from the bandage prototype tended to decrease as time passed.

In terms of kinematics, both prototypes resulted in a reduced step length: the median of the average \SI{0.58}{m} at baseline, compared to \SI{0.53}{m} with the cuff and \SI{0.53}{m} with the bandage. Step height also changed with the system: the median of the average at baseline was \SI{0.16}{m}, compared to the cuff condition at \SI{0.15}{m} and the bandage at \SI{0.10}{m}.

\begin{figure}[t]
    \centering
   \def\svgwidth{1\linewidth}
    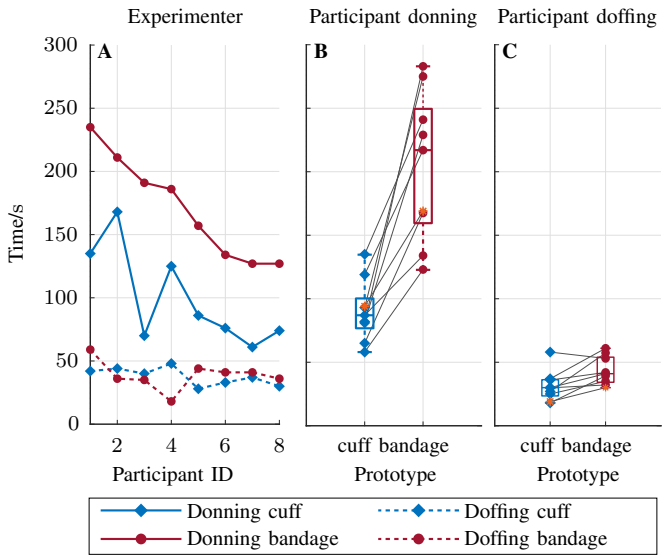
    \caption{Donning and doffing time of both prototypes. \textbf{A}: Donning (solid) and doffing (dashed) by the experimenter over time for the cuff (blue diamond) and bandage (red circle) prototype. Donning (\textbf{B}) and doffing (\textbf{C}) of the two prototypes by the participants and therapist (orange star) are also shown.}
    \label{fig:results_don}
\end{figure}

Donning and doffing for both prototypes took under five minutes (Fig.~\ref{fig:results_don}). Experimenter donning times improved with practice: cuff prototype times decreased from \SI{135}{s} to \SI{74}{s}, and bandage times from \SI{235}{s} to \SI{127}{s} (Fig.~\ref{fig:results_don}A). Doffing times remained consistent at \qtyrange{30}{40}{s}. Participants averaged \SI{87}{s} to don the cuff and \SI{217}{s} for the bandage (Fig.~\ref{fig:results_don}B), with doffing averaging \SI{30}{s} for the cuff and \SI{41}{s} for the bandage (Fig.~\ref{fig:results_don}C).

\begin{figure}[t]
    \centering
    \def\svgwidth{1\linewidth}
    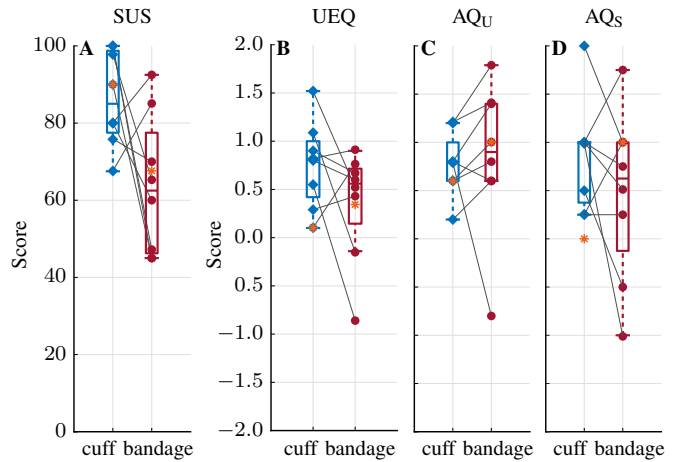
    \caption{Results of the questionnaires for the cuff (blue diamond) and bandage (red circle) prototypes. Therapist scores are shown as orange stars. The System Usability Scale (SUS) (\textbf{A}) is ranked on a scale from 0 to 100, and the User Experience Questionnaire (UEQ) (\textbf{B}), Acceptance Questionnaire on Usability (AS\textsubscript{U}) (\textbf{C}) and Satisfaction (AS\textsubscript{S}) (\textbf{D}) on a scale from -2 to 2.}
    \label{fig:res_quest}
\end{figure}

The questionnaire results are presented in Fig.~\ref{fig:res_quest}. Both prototypes received generally positive feedback on user experience (UEQ), with median scores of 0.8 for the cuff and 0.6 for the bandage prototype, indicating a more favorable user experience with the cuff. Based on the responses on usability (SUS), the cuff prototype was rated as highly usable, scoring 85 out of 100, while the bandage prototype scored lower at 62.5 out of 100. The P-MATE as a system, as well as the two human-device interfaces, were scored on usefulness and satisfaction on the Van der Laan Acceptance Questionnaire. Similar to the other questionnaires, both prototypes received positive ratings on both usefulness and satisfaction, with the cuff prototype rated higher than the bandage prototype.

In the study, one participant lost balance within the first minute of using the bandage prototype. Analysis of the kinematics indicated that the participant was taking longer steps just before the loss of balance, generating a strong forward pulling force from the elastic tendons (\SI{235.6}{N}), causing instability in the subsequent step. Following this incident, the study was temporarily paused, and the participant restarted with an elongated anterior tendon and a reduced treadmill speed. No further incidents occurred for the remainder of the study.

While the therapist liked the cuff prototype less than the bandage, both were preferred over a rigid exoskeleton due to their greater freedom of movement and more natural gait. Walking in both prototypes became ``easier with practice,'' though he ``needed to adjust the gait timing'' with the cuff prototype to align with the P-MATE. He also recommended moving the front attachment of the bandage prototype closer to the toes, noting that the lengthened elastic tendon provided ``more guidance than control'' over foot placement.

%% file: 0T_Questionnaire.tex
\begin{table}
\centering
\caption{Demographic data of experiment participants and their chosen treadmill walking speed. T: Therapist}
\begin{tabular}{llllll}
\toprule
\textbf{ID} &\textbf{Sex}  & \textbf{Age/\si{yr}}  & \textbf{Height/\si{m}} & \textbf{Mass/\si{kg}}  & \textbf{Speed/\si{km/h}} \\ \midrule
1           & F        & 63 & 1.60          & 79      &  3.0              \\ 
2           & F        & 23 & 1.65          & 70      &  3.0              \\ 
3           & M        & 26 & 1.85          & 72      &  2.5              \\ 
4           & M        & 61 & 1.73          & 79      &  3.5              \\
5           & M        & 22 & 1.86          & 94      &  3.5              \\
6           & M        & 24 & 1.91          & 86      &  3.5            \\
7           & F        & 24 & 1.67          & 70      &  3.0              \\
8           & F        & 60 & 1.63          & 71      &  2.5              \\
T   & M        & 49 & 1.81          & 73      &  3.0             \\
\bottomrule
\end{tabular}
\label{tab:exp_subjects}
\end{table}

%% file: F3_setup.pdf_tex
\begingroup%
  \makeatletter%
  \providecommand\color[2][]{%
    \errmessage{(Inkscape) Color is used for the text in Inkscape, but the package 'color.sty' is not loaded}%
    \renewcommand\color[2][]{}%
  }%
  \providecommand\transparent[1]{%
    \errmessage{(Inkscape) Transparency is used (non-zero) for the text in Inkscape, but the package 'transparent.sty' is not loaded}%
    \renewcommand\transparent[1]{}%
  }%
  \providecommand\rotatebox[2]{#2}%
  \newcommand*\fsize{\dimexpr\f@size pt\relax}%
  \newcommand*\lineheight[1]{\fontsize{\fsize}{#1\fsize}\selectfont}%
  \ifx\svgwidth\undefined%
    \setlength{\unitlength}{525bp}%
    \ifx\svgscale\undefined%
      \relax%
    \else%
      \setlength{\unitlength}{\unitlength * \real{\svgscale}}%
    \fi%
  \else%
    \setlength{\unitlength}{\svgwidth}%
  \fi%
  \global\let\svgwidth\undefined%
  \global\let\svgscale\undefined%
  \makeatother%
  \begin{picture}(1,0.85714286)%
    \lineheight{1}%
    \setlength\tabcolsep{0pt}%
    \put(0,0){\includegraphics[width=\unitlength,page=1]{F3_setup.pdf}}%
    \put(0.25207784,0.12524929){\color[rgb]{0,0,0}\makebox(0,0)[t]{\lineheight{1.25}\smash{\begin{tabular}[t]{c}\footnotesize{Participant ID}\end{tabular}}}}%
    \put(0.4998419,0.16872076){\color[rgb]{0,0,0}\makebox(0,0)[lt]{\lineheight{1.25}\smash{\begin{tabular}[t]{l}\footnotesize{cuff bandage}\end{tabular}}}}%
    \put(0.27079792,0.81294146){\color[rgb]{0,0,0}\makebox(0,0)[t]{\lineheight{0.80000001}\smash{\begin{tabular}[t]{c}\footnotesize{Experimenter}\\.\end{tabular}}}}%
    \put(0.16846319,0.16861606){\color[rgb]{0,0,0}\makebox(0,0)[t]{\lineheight{1.25}\smash{\begin{tabular}[t]{c}\footnotesize{$2$}\end{tabular}}}}%
    \put(0.24992291,0.16907737){\color[rgb]{0,0,0}\makebox(0,0)[t]{\lineheight{1.25}\smash{\begin{tabular}[t]{c}\footnotesize{$4$}\end{tabular}}}}%
    \put(0.33065522,0.16866071){\color[rgb]{0,0,0}\makebox(0,0)[t]{\lineheight{1.25}\smash{\begin{tabular}[t]{c}\footnotesize{$6$}\end{tabular}}}}%
    \put(0.41193631,0.16857142){\color[rgb]{0,0,0}\makebox(0,0)[t]{\lineheight{1.25}\smash{\begin{tabular}[t]{c}\footnotesize{$8$}\end{tabular}}}}%
    \put(0.02685361,0.4974081){\color[rgb]{0,0,0}\rotatebox{90}{\makebox(0,0)[t]{\lineheight{1.25}\smash{\begin{tabular}[t]{c}\footnotesize{Time/\si{s}}\end{tabular}}}}}%
    \put(0.7741276,0.16872076){\color[rgb]{0,0,0}\makebox(0,0)[lt]{\lineheight{1.25}\smash{\begin{tabular}[t]{l}\footnotesize{cuff bandage}\end{tabular}}}}%
    \put(0.10312895,0.20191269){\color[rgb]{0,0,0}\makebox(0,0)[t]{\lineheight{1.25}\smash{\begin{tabular}[t]{c}\footnotesize{$0$}\end{tabular}}}}%
    \put(0.09344145,0.29683333){\color[rgb]{0,0,0}\makebox(0,0)[t]{\lineheight{1.25}\smash{\begin{tabular}[t]{c}\footnotesize{$50$}\end{tabular}}}}%
    \put(0.08375395,0.39175396){\color[rgb]{0,0,0}\makebox(0,0)[t]{\lineheight{1.25}\smash{\begin{tabular}[t]{c}\footnotesize{$100$}\end{tabular}}}}%
    \put(0.08375395,0.4866746){\color[rgb]{0,0,0}\makebox(0,0)[t]{\lineheight{1.25}\smash{\begin{tabular}[t]{c}\footnotesize{$150$}\end{tabular}}}}%
    \put(0.08375395,0.58159522){\color[rgb]{0,0,0}\makebox(0,0)[t]{\lineheight{1.25}\smash{\begin{tabular}[t]{c}\footnotesize{$200$}\end{tabular}}}}%
    \put(0.08375395,0.67651587){\color[rgb]{0,0,0}\makebox(0,0)[t]{\lineheight{1.25}\smash{\begin{tabular}[t]{c}\footnotesize{$250$}\end{tabular}}}}%
    \put(0.08375395,0.7714365){\color[rgb]{0,0,0}\makebox(0,0)[t]{\lineheight{1.25}\smash{\begin{tabular}[t]{c}\footnotesize{$300$}\end{tabular}}}}%
    \put(0.85947915,0.81340277){\color[rgb]{0,0,0}\makebox(0,0)[t]{\lineheight{0.80000001}\smash{\begin{tabular}[t]{c}\footnotesize{Participant} \footnotesize{doffing}\end{tabular}}}}%
    \put(0.52458144,0.12524929){\color[rgb]{0,0,0}\makebox(0,0)[lt]{\lineheight{1.25}\smash{\begin{tabular}[t]{l}\footnotesize{Prototype}\end{tabular}}}}%
    \put(0.79886715,0.12524929){\color[rgb]{0,0,0}\makebox(0,0)[lt]{\lineheight{1.25}\smash{\begin{tabular}[t]{l}\footnotesize{Prototype}\end{tabular}}}}%
    \put(0,0){\includegraphics[width=\unitlength,page=2]{F3_setup.pdf}}%
    \put(0.27331219,0.0699772){\color[rgb]{0,0,0}\makebox(0,0)[lt]{\lineheight{1.25}\smash{\begin{tabular}[t]{l}\footnotesize{Donning cuff}\end{tabular}}}}%
    \put(0.68974077,0.02998541){\color[rgb]{0,0,0}\makebox(0,0)[lt]{\lineheight{1.25}\smash{\begin{tabular}[t]{l}\footnotesize{Doffing bandage}\end{tabular}}}}%
    \put(0.68974077,0.0699772){\color[rgb]{0,0,0}\makebox(0,0)[lt]{\lineheight{1.25}\smash{\begin{tabular}[t]{l}\footnotesize{Doffing cuff}\end{tabular}}}}%
    \put(0.27331219,0.03005981){\color[rgb]{0,0,0}\makebox(0,0)[lt]{\lineheight{1.25}\smash{\begin{tabular}[t]{l}\footnotesize{Donning bandage}\end{tabular}}}}%
    \put(0,0){\includegraphics[width=\unitlength,page=3]{F3_setup.pdf}}%
    \put(0.58328868,0.81340277){\color[rgb]{0,0,0}\makebox(0,0)[t]{\lineheight{0.80000001}\smash{\begin{tabular}[t]{c}\footnotesize{Participant} \footnotesize{donning}\end{tabular}}}}%
    \put(0,0){\includegraphics[width=\unitlength,page=4]{F3_setup.pdf}}%
    \put(0.1504993,0.76138935){\color[rgb]{0,0,0}\makebox(0,0)[t]{\lineheight{1.25}\smash{\begin{tabular}[t]{c}\footnotesize{\textbf{A}}\end{tabular}}}}%
    \put(0.7578338,0.76138935){\color[rgb]{0,0,0}\makebox(0,0)[t]{\lineheight{1.25}\smash{\begin{tabular}[t]{c}\footnotesize{\textbf{C}}\end{tabular}}}}%
    \put(0.47478322,0.76138935){\color[rgb]{0,0,0}\makebox(0,0)[t]{\lineheight{1.25}\smash{\begin{tabular}[t]{c}\footnotesize{\textbf{B}}\end{tabular}}}}%
    \put(0,0){\includegraphics[width=\unitlength,page=5]{F3_setup.pdf}}%
  \end{picture}%
\endgroup%

%% file: F4_Quest.pdf_tex
\begingroup%
  \makeatletter%
  \providecommand\color[2][]{%
    \errmessage{(Inkscape) Color is used for the text in Inkscape, but the package 'color.sty' is not loaded}%
    \renewcommand\color[2][]{}%
  }%
  \providecommand\transparent[1]{%
    \errmessage{(Inkscape) Transparency is used (non-zero) for the text in Inkscape, but the package 'transparent.sty' is not loaded}%
    \renewcommand\transparent[1]{}%
  }%
  \providecommand\rotatebox[2]{#2}%
  \newcommand*\fsize{\dimexpr\f@size pt\relax}%
  \newcommand*\lineheight[1]{\fontsize{\fsize}{#1\fsize}\selectfont}%
  \ifx\svgwidth\undefined%
    \setlength{\unitlength}{525bp}%
    \ifx\svgscale\undefined%
      \relax%
    \else%
      \setlength{\unitlength}{\unitlength * \real{\svgscale}}%
    \fi%
  \else%
    \setlength{\unitlength}{\svgwidth}%
  \fi%
  \global\let\svgwidth\undefined%
  \global\let\svgscale\undefined%
  \makeatother%
  \begin{picture}(1,0.71428571)%
    \lineheight{1}%
    \setlength\tabcolsep{0pt}%
    \put(0,0){\includegraphics[width=\unitlength,page=1]{F4_Quest.pdf}}%
    \put(0.12292833,0.608848){\color[rgb]{0,0,0}\makebox(0,0)[t]{\lineheight{1.25}\smash{\begin{tabular}[t]{c}\footnotesize{\textbf{A}}\end{tabular}}}}%
    \put(0.63584982,0.608848){\color[rgb]{0,0,0}\makebox(0,0)[t]{\lineheight{1.25}\smash{\begin{tabular}[t]{c}\footnotesize{\textbf{C}}\end{tabular}}}}%
    \put(0.41808053,0.608848){\color[rgb]{0,0,0}\makebox(0,0)[t]{\lineheight{1.25}\smash{\begin{tabular}[t]{c}\footnotesize{\textbf{B}}\end{tabular}}}}%
    \put(0.83257382,0.608848){\color[rgb]{0,0,0}\makebox(0,0)[t]{\lineheight{1.25}\smash{\begin{tabular}[t]{c}\footnotesize{\textbf{D}}\end{tabular}}}}%
    \put(0.1997578,0.00857309){\color[rgb]{0,0,0}\makebox(0,0)[t]{\lineheight{1.10000002}\smash{\begin{tabular}[t]{c}\footnotesize{cuff bandage}\end{tabular}}}}%
    \put(0.90707624,0.00857309){\color[rgb]{0,0,0}\makebox(0,0)[t]{\lineheight{1.10000002}\smash{\begin{tabular}[t]{c}\footnotesize{cuff bandage}\end{tabular}}}}%
    \put(0.50272962,0.00857309){\color[rgb]{0,0,0}\makebox(0,0)[t]{\lineheight{1.10000002}\smash{\begin{tabular}[t]{c}\footnotesize{cuff bandage}\end{tabular}}}}%
    \put(0.70981454,0.00857309){\color[rgb]{0,0,0}\makebox(0,0)[t]{\lineheight{1.10000002}\smash{\begin{tabular}[t]{c}\footnotesize{cuff bandage}\end{tabular}}}}%
    \put(0.38793762,0.03731641){\color[rgb]{0,0,0}\makebox(0,0)[rt]{\lineheight{1.10000002}\smash{\begin{tabular}[t]{r}\footnotesize{$-2.0$}\end{tabular}}}}%
    \put(0.38793766,0.54443697){\color[rgb]{0,0,0}\makebox(0,0)[rt]{\lineheight{1.10000002}\smash{\begin{tabular}[t]{r}\footnotesize{$1.5$}\end{tabular}}}}%
    \put(0.38793766,0.47199118){\color[rgb]{0,0,0}\makebox(0,0)[rt]{\lineheight{1.10000002}\smash{\begin{tabular}[t]{r}\footnotesize{$1.0$}\end{tabular}}}}%
    \put(0.38793766,0.39954538){\color[rgb]{0,0,0}\makebox(0,0)[rt]{\lineheight{1.10000002}\smash{\begin{tabular}[t]{r}\footnotesize{$0.5$}\end{tabular}}}}%
    \put(0.38793766,0.32709961){\color[rgb]{0,0,0}\makebox(0,0)[rt]{\lineheight{1.10000002}\smash{\begin{tabular}[t]{r}\footnotesize{$0.0$}\end{tabular}}}}%
    \put(0.38793766,0.2546538){\color[rgb]{0,0,0}\makebox(0,0)[rt]{\lineheight{1.10000002}\smash{\begin{tabular}[t]{r}\footnotesize{$0.5$}\end{tabular}}}}%
    \put(0.38793762,0.18220799){\color[rgb]{0,0,0}\makebox(0,0)[rt]{\lineheight{1.10000002}\smash{\begin{tabular}[t]{r}\footnotesize{$-1.0$}\end{tabular}}}}%
    \put(0.38793762,0.10976222){\color[rgb]{0,0,0}\makebox(0,0)[rt]{\lineheight{1.10000002}\smash{\begin{tabular}[t]{r}\footnotesize{$-1.5$}\end{tabular}}}}%
    \put(0.38793766,0.61688276){\color[rgb]{0,0,0}\makebox(0,0)[rt]{\lineheight{1.10000002}\smash{\begin{tabular}[t]{r}\footnotesize{$2.0$}\end{tabular}}}}%
    \put(0.09252889,0.03556549){\color[rgb]{0,0,0}\makebox(0,0)[rt]{\lineheight{1.10000002}\smash{\begin{tabular}[t]{r}\footnotesize{$0$}\end{tabular}}}}%
    \put(0.09252887,0.38330532){\color[rgb]{0,0,0}\makebox(0,0)[rt]{\lineheight{1.10000002}\smash{\begin{tabular}[t]{r}\footnotesize{$60$}\end{tabular}}}}%
    \put(0.09252887,0.26739205){\color[rgb]{0,0,0}\makebox(0,0)[rt]{\lineheight{1.10000002}\smash{\begin{tabular}[t]{r}\footnotesize{$40$}\end{tabular}}}}%
    \put(0.09252887,0.15147875){\color[rgb]{0,0,0}\makebox(0,0)[rt]{\lineheight{1.10000002}\smash{\begin{tabular}[t]{r}\footnotesize{$20$}\end{tabular}}}}%
    \put(0.09252888,0.61513185){\color[rgb]{0,0,0}\makebox(0,0)[rt]{\lineheight{1.10000002}\smash{\begin{tabular}[t]{r}\footnotesize{$100$}\end{tabular}}}}%
    \put(0.09252887,0.49921858){\color[rgb]{0,0,0}\makebox(0,0)[rt]{\lineheight{1.10000002}\smash{\begin{tabular}[t]{r}\footnotesize{$80$}\end{tabular}}}}%
    \put(0.90032027,0.65653977){\color[rgb]{0,0,0}\makebox(0,0)[t]{\lineheight{1.10000002}\smash{\begin{tabular}[t]{c}\footnotesize{AQ\textsubscript{S}}\end{tabular}}}}%
    \put(0.70305856,0.65653977){\color[rgb]{0,0,0}\makebox(0,0)[t]{\lineheight{1.10000002}\smash{\begin{tabular}[t]{c}\footnotesize{AQ\textsubscript{U}}\end{tabular}}}}%
    \put(0.49411354,0.65653977){\color[rgb]{0,0,0}\makebox(0,0)[t]{\lineheight{1.10000002}\smash{\begin{tabular}[t]{c}\footnotesize{UEQ}\end{tabular}}}}%
    \put(0,0){\includegraphics[width=\unitlength,page=2]{F4_Quest.pdf}}%
    \put(0.19177419,0.65653977){\color[rgb]{0,0,0}\makebox(0,0)[t]{\lineheight{1.10000002}\smash{\begin{tabular}[t]{c}\footnotesize{SUS}\end{tabular}}}}%
    \put(0.03177976,0.32419791){\color[rgb]{0,0,0}\rotatebox{90}{\makebox(0,0)[t]{\lineheight{1.10000002}\smash{\begin{tabular}[t]{c}\footnotesize{Score}\end{tabular}}}}}%
    \put(0.3239226,0.32419791){\color[rgb]{0,0,0}\rotatebox{90}{\makebox(0,0)[t]{\lineheight{1.10000002}\smash{\begin{tabular}[t]{c}\footnotesize{Score}\end{tabular}}}}}%
    \put(0,0){\includegraphics[width=\unitlength,page=3]{F4_Quest.pdf}}%
  \end{picture}%
\endgroup%

%% file: 04_Discussion.tex
\section{Discussion} 
In this study, we designed, built, and evaluated two prototypes of the Passive Mechanical Add-on for Treadmill Exercise (P-MATE). This device could assist stroke patients during their gait rehabilitation by leveraging the non-paretic leg to facilitate the swing phase of the paretic leg using an elastic tendon-and-pulley system. The system was optimized in simulation to minimize parasitic forces acting on the user during natural treadmill walking. 

Results from a feasibility study with eight adults without physical impairments and a case study with one physical therapist indicate differences in user experience between the two human-device interfaces. Higher anterior pulling forces were measured with the bandage prototype compared to the cuff interface. This aligns with the therapist's observation of ``feeling more guided'' using the cuff and a ``more controlling'' perception when wearing the bandage prototype.
Some design elements of the current setup require further improvement. A general solution for the location of the pulleys was applied for all participants. However, variations in user experience, pretension, and walking speed preferences suggest that incorporating user-specific configurations could improve the experience of using the P-MATE for gait rehabilitation. Additionally, the pulleys have a fixed orientation, which prevents dynamic adjustments of the guiding of elastic tendons throughout the gait cycle. This limitation may cause friction and wear on the tendons, potentially affecting both their durability and the user’s kinematics.

The initial pretension differed per participant, influencing the interaction forces of the P-MATE. In the case of the participant who lost their balance and the therapist, the tendons were lengthened to reduce the pretension. Both the participant and therapist reported feeling in more control with the longer tendons compared to the original tendon length, possibly as a result of the reduced
forces from the device. All participants claimed that walking became easier with practice, particularly when using the bandage prototype, in line with the observed reduction in the interaction forces. A future system should allow for rapid customization of the pulley locations and pretension of the elastic tendons to maximize the clinical impact of the system.

While the depth camera allowed for a quick and cheap data collection, we likely need to optimize the setup and use different sensors to collect more meaningful data. First, the current experimental setup only collected kinematic data from the right leg, limiting insights into the full impact of the P-MATE prototypes on the overall gait pattern. 
Secondly, we noticed that the ankle marker shifted throughout the various walking sessions, leading to an inaccurate indication of the anatomical ankle marker. Despite these limitations, the kinematic data indicates a reduced step length for both prototypes. In addition to limitations in the kinematic measurements, it is important to note the two load cells used in this study measured the combined forces from the two anterior and two posterior elastic tendons, respectively, making it impossible to distinguish the individual pulling force contributions from each tendon acting on the leg.  Participant feedback suggested also biomechanical changes that could not be validated by our dataset, including perceiving changes in balance and energy expenditure, so relevant sensors should be added to the future protocols.

Further research is needed to better understand the P-MATE’s impact on the gait of people who had a stroke. Insights from these experiments would help assess whether P-MATE promotes healthy muscle engagement, which is essential for motor learning and rehabilitation while ensuring active participation from users. Additionally, the long-term effects of the P-MATE on both the quantity of training and quality of movement should be considered to ensure that the P-MATE does not lead to negative outcomes, such as over-reliance on the device or compensatory movement patterns that could hinder recovery.

%% file: 04b_Conclusion.tex
\section{Conclusion}
We presented the design and evaluation of a passive mechanical treadmill add-on, the P-MATE, for gait rehabilitation following a stroke. The system employs elastic tendons, pulleys, and treadmill actuation to create a dynamic connection between legs, in order to leverage the movement of the non-paretic leg to assist the paretic limb. The P-MATE was evaluated on kinematics and forces alongside unimpaired participants' feedback on the system acceptance, usability, and user experience.

Although differences were observed between normal treadmill walking kinematics and walking with both P-MATE human-device interfaces, participants responded positively regarding the usability and overall experience with the device, showing promise toward accessible gait therapy. However, several design aspects require refinement, including addressing pulley derailing, optimizing tendon pretension, and conducting a more detailed kinematic analysis. Future work will also focus on further improving usability and clinical impact.